\documentclass[10pt,twocolumn,letterpaper]{article}

\usepackage[pagenumbers]{cvpr} 

\usepackage[pagebackref,breaklinks,colorlinks,citecolor=cvprblue]{hyperref}
\usepackage{multirow}
\usepackage{makecell}
\usepackage{bookmark}
\usepackage{algorithm} 
\usepackage{listings}
\usepackage{arydshln}

\definecolor{cvprblue}{rgb}{0.21,0.49,0.74}
\definecolor{codeblue}{rgb}{0.25,0.5,0.5}
\definecolor{codekw}{rgb}{0.85, 0.18, 0.50}
\definecolor{std}{rgb}{0.7529,0.4902,0.6471}

\lstdefinestyle{Python}{
  language=python,
  backgroundcolor=\color{white},
  basicstyle=\fontsize{7.5pt}{7.5pt}\ttfamily\selectfont,
  columns=fullflexible,
  breaklines=true,
  captionpos=b,
  commentstyle=\fontsize{7.5pt}{7.5pt}\color{codeblue},
  keywordstyle=\fontsize{7.5pt}{7.5pt}\color{codekw},
  morekeywords={assert},
}

\newcommand{\valstd}[2]{$#1 \textcolor{std}{\scriptstyle \;\pm\, #2}$}
\newcommand{\valstdb}[2]{$\mathbf{#1} \textcolor{std}{\scriptstyle \,\pm\, #2}$}
 \newcommand{\modelname}{RepNeXt}

\def\paperID{*****} 
\def\confName{CVPR}
\def\confYear{2024}

\title{\modelname: A Fast Multi-Scale CNN using Structural Reparameterization}

\author{Mingshu Zhao$^{1}$ \quad Yi Luo$^{1}$ \quad Yong Ouyang$^{1,2}$  \\
$^1$Sichuan Energy Internet Research Institute, Tsinghua University \\
$^2$Chengdu Qingrong Shentong Technology \\
{\tt\small zhaomingshu@tsinghua-eiri.org \quad luoyi@tsinghua-eiri.org \quad ouyangyong@deepsensing.cn}}

\begin{document}
\maketitle
\begin{abstract}
    In the realm of resource-constrained mobile vision tasks, the pursuit of efficiency and performance consistently drives innovation in lightweight Convolutional Neural Networks (CNNs) and Vision Transformers (ViTs).
    While ViTs excel at capturing global context through self-attention mechanisms, their deployment in resource-limited environments is hindered by computational complexity and latency.
    Conversely, lightweight CNNs are favored for their parameter efficiency and low latency. 
    This study investigates the complementary advantages of CNNs and ViTs to develop a versatile vision backbone tailored for resource-constrained applications.
    We introduce \modelname{}, a novel model series integrates multi-scale feature representations and incorporates both serial and parallel structural reparameterization (SRP) to enhance network depth and width without compromising inference speed.
    Extensive experiments demonstrate \modelname{}'s superiority over current leading lightweight CNNs and ViTs, providing advantageous latency across various vision benchmarks.
    \modelname{}-M4 matches RepViT-M1.5's 82.3\% accuracy on ImageNet within 1.5ms on an iPhone 12, outperforms its AP$^{box}$ by 1.3 on MS-COCO, and reduces parameters by 0.7M.
    Codes and models are available at \url{https://github.com/suous/RepNeXt}.
\end{abstract}
    
\section{Introduction}
\label{sec:intro}

Over the past decade, Convolutional Neural Networks (CNNs)~\cite{krizhevsky2012imagenet,he2016deep,redmon2016you} have been predominant in computer vision applications, leveraging their inherent locality and translation equivariance~\cite{dosovitskiy2020image}.
To facilitate their deployment on resource-constrained devices, various efficient design principles have emerged, including spatial or depth separable convolutions~\cite{szegedy2016rethinking,szegedy2017inception,howard2017mobilenets},
channel shuffling~\cite{zhang2018shufflenet},
partial channel operations~\cite{ghostnet,FasterNet,SHVIT},
neural architecture search~\cite{tan2019mnasnet,tan2019efficientnet},
network pruning~\cite{han2015deep_compression,Wen_NIPS2016,wang2021convolutional},
and structural reparameterization (SRP)~\cite{ding2021repvgg,ding2019acnet}.

\begin{figure}[t]
  \centering
    \includegraphics[width=1.0\linewidth]{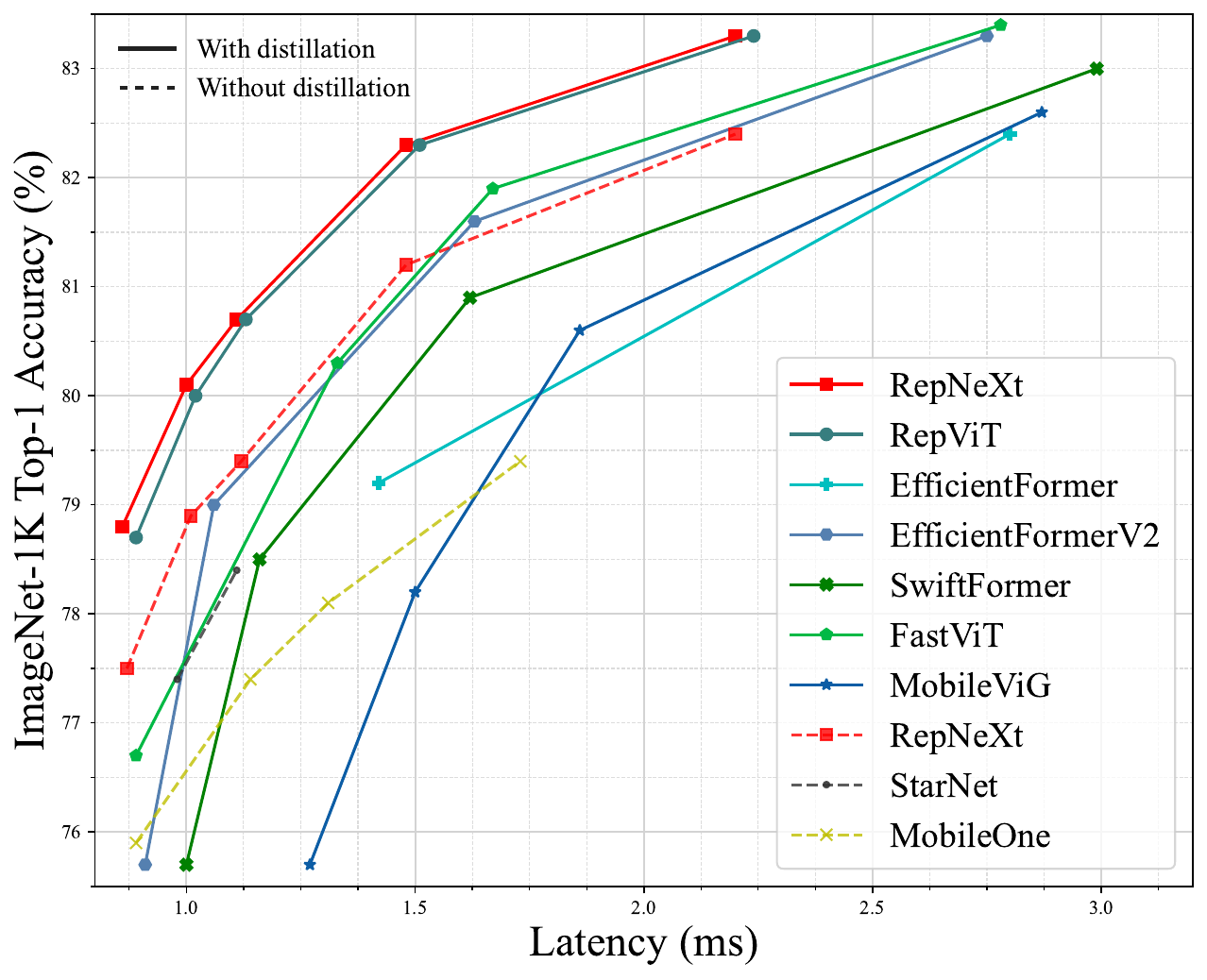}
    \caption{\textbf{Latency vs Accuracy Comparison}. The top-1 accuracy is tested on ImageNet-1K and the latency is measured by an iPhone 12 with iOS 16 across 20 experimental sets. 
    \modelname{} consistently achieves the best trade-off between performance and latency.}
    \label{fig:introduction}
    \vspace{-0.5cm}
\end{figure}

Vision Transformers (ViTs)~\cite{dosovitskiy2020image,carion2020end} have emerged as a competitive alternative to CNNs, with several innovations aimed at improving their efficiency, such as
hierarchical designs or hybrid architectures~\cite{wang2021pyramid,li2021improved,dai2021coatnet,xia2024vit,hatamizadeh2023fastervit},
as well as local processing operations or linear attention mechanisms~\cite{hassani2023neighborhood,pan2023slide,dong2022cswin,han2023flatten,cai2022efficientvit}.
However, many optimizations require special operations that may not be feasible on devices with limited resources.
Meanwhile, efficient designs often prioritize optimizing inference speed based on metrics like floating point operations or model sizes, which may not consistently correlate with actual latency experienced in mobile applications.
Consequently, convolution operations are still preferred for balancing latency and accuracy~\cite{wang2023repvit}.

Inspired by the sophisticated architectures~\cite{yu2022metaformer} of ViTs and their ability to model long-range spatial dependencies~\cite{dosovitskiy2020image}, large-kernel CNNs~\cite{liu2022convnet,ding2022scaling,liu2022more} have gained widespread research attention for enlarging the effective receptive field (ERF).
However, expanding kernel sizes may substantially inflate parameter counts, resulting in considerable memory requirements and optimization challenges.

To balance performance and speed while preserving both local and global representations, we present \modelname{}, a multi-scale CNN inspired by MixConv~\cite{tan2019mixconv} and InceptionNeXt~\cite{yu2023inceptionnext}.
\modelname{} combines the hierarchical design of CNNs~\cite{krizhevsky2012imagenet,he2016deep} with the general architecture of ViTs~\cite{vaswani2017attention,yu2022metaformer,liu2022convnet,yu2023inceptionnext} at a macro level, and integrates the efficiency of small-kernel convolutions with the broad perspective of large-kernel convolutions at a micro level.
Extensive experiments demonstrate its effectiveness across various vision benchmarks, including ImageNet-1K~\cite{deng2009imagenet} for image classification, MS-COCO~\cite{lin2014microsoft} for object detection and instance segmentation, and ADE20K~\cite{zhou2017scene} for semantic segmentation.
Our contributions can be concluded as follows.
\begin{itemize}[leftmargin=1em]
\setlength
\itemsep{0em}
    \item We introduce \modelname{}, a simple yet effective vision backbone with a consistent design across inner-stage blocks and downsampling layers, achieving competitive or superior performance considering the trade-off between accuracy and latency with only fundamental operation units, facilitating subsequent optimizations.
    \item We leverage both serial and parallel SRP mechanisms to increase network depth and width during training, effectively improving representational capacity without sacrificing inference speed.
    \item Following~\cite{wang2023repvit}, we further demonstrate that a simple multi-scale CNN (without channel attention blocks~\cite{hu2018squeeze}) can outperform sophisticated architectures or complicated operators through intricate design or neural architecture search (NAS) across various vision tasks.
\end{itemize}

\section{Related Work}

\noindent\textbf{Efficient CNNs}: Crafting efficient CNNs for edge vision applications has received a lot of attention in recent years.
MobileNets~\cite{howard2017mobilenets,sandler2018mobilenetv2,howard2019searching} proposed depthwise separable convolutions as well as inverted residual blocks for better efficiency-accuracy trade-off.
SqueezeNet~\cite{SqueezeNet} used squeeze and expand operations to maintain representational capacity while reducing computational cost.
ShuffleNet~\cite{zhang2018shufflenet} implemented channel shuffle after pointwise group convolutions for improved information exchange.
GhostNet~\cite{ghostnet} and FasterNet~\cite{FasterNet} introduced partial channel operations to generate feature maps more efficiently.
MicroNet~\cite{li2021micronet} aggressively reduced FLOPs through further network decomposition and sparsification.
MnasNet~\cite{tan2019mnasnet} and EfficientNet~\cite{tan2019efficientnet} leveraged neural architecture search (NAS) to automatically discover efficient architectures.
ParC-Net~\cite{zhang2022parcnet} proposed position-aware circular convolution (ParC) to provide a global receptive field while producing location-sensitive features.
StarNet~\cite{ma2024rewrite}  demonstrated the efficacy of star operation in extracting substantial representation power from implicitly high-dimensional spaces.
RepViT~\cite{wang2023repvit} integrated architectural designs from efficient ViTs into mobile CNNs, leveraging SRP~\cite{ding2019acnet,ding2021repvgg} techniques and SE~\cite{hu2018squeeze} modules to boost performance.
Furthermore, network pruning~\cite{han2015deep_compression,Wen_NIPS2016,wang2021convolutional} and low-bit quantization~\cite{wu2016quantized} mechanisms are often employed to further reduce model size and memory usage.

\noindent\textbf{Efficient ViTs}: Recent advancements in efficient ViTs concentrate on incorporating spatial inductive biases within ViT blocks.
MobileViTs~\cite{mehta2021mobilevit,mehta2022separable,wadekar2022mobilevitv3} integrated the efficiency of MobileNets with the global modeling capabilities of ViTs.
Mobile-Former~\cite{chen2022mobile} utilized a bidirectional parallel structure to facilitate interaction between local and global features.
EfficientFormers~\cite{li2022efficientformer,li2022rethinking} featured a dimension-consistent design using hardware-friendly 4D modules and powerful 3D Multi-Head Self-Attention (MHSA) blocks.
FastViT~\cite{vasu2023fastvit} combined $7\times7$ depthwise convolutions and SRP to improve model capacity and efficiency.
EdgeViTs~\cite{pan2022edgevits} innovated with Local-Global-Local blocks to better integrate MHSA and convolution.
SwiftFormer~\cite{Shaker_2023_ICCV} introduced an efficient additive attention mechanism that replaces the quadratic matrix multiplications with linear element-wise operations.
LightViT~\cite{huang2022lightvit} incorporated a global aggregation scheme into both token and channel mixers to achieve a superior performance-efficiency trade-off.
SHVIT~\cite{SHVIT} addressed computational redundancies with a Single-Head Self-Attention (SHSA) on subset of channels.

\begin{figure*}[t]
\centering
    \includegraphics[width=0.95\linewidth]{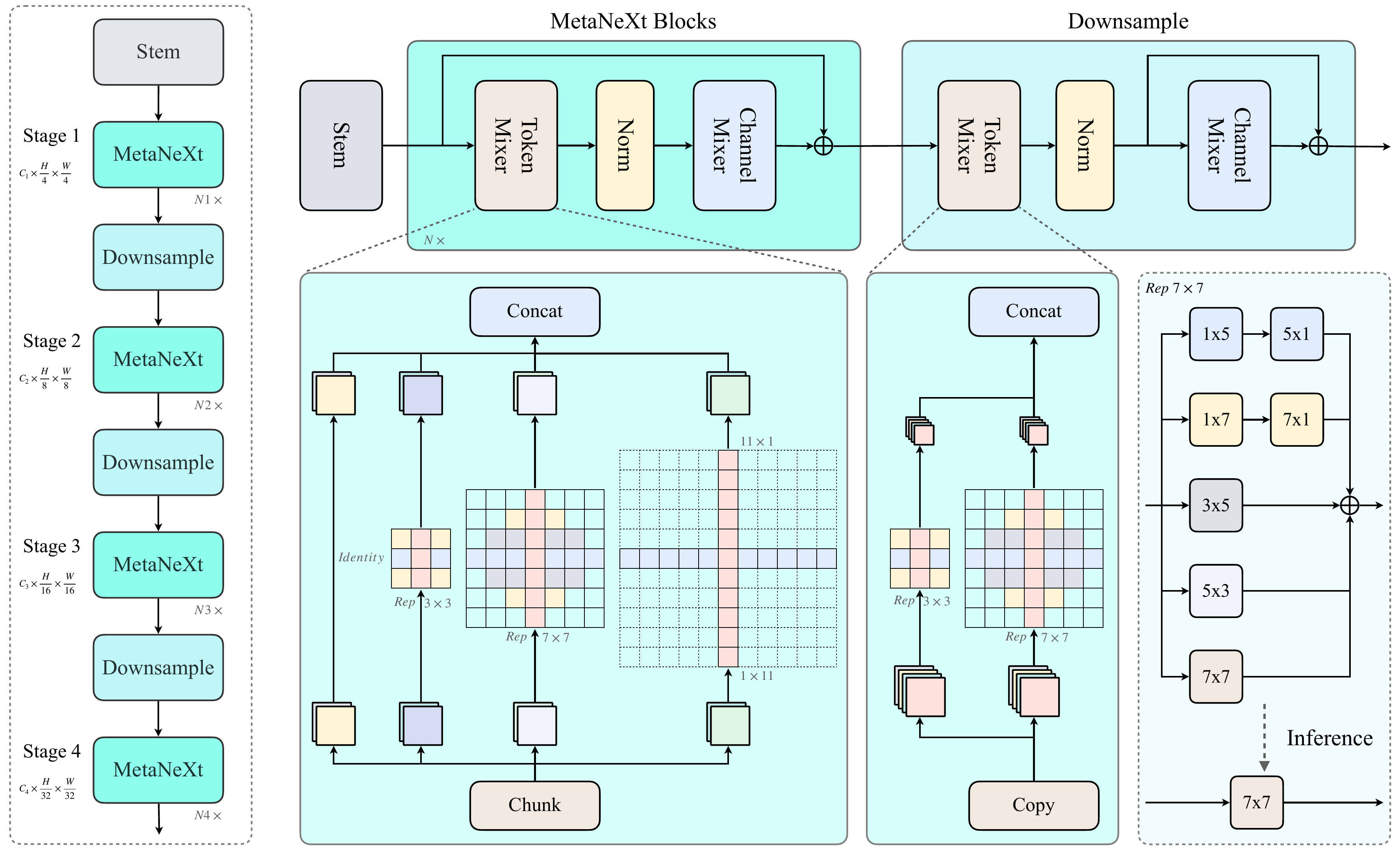}
    \caption{(left) \textbf{The macro architecture of \modelname{}.} \modelname{} adopts a four-stage hierarchical design, starting with two $3\times3$ convolutions with a stride of $2$. Where $C_i$ represent channel dimensions at stage $i$, while $H$ and $W$ denote image height and width, respectively. (right) \textbf{The micro design of MetaNeXt and Downsampling blocks.} 
    The MetaNeXt block~\cite{liu2022convnet,yu2023inceptionnext} includes a \textit{token mixer} for spatial feature extraction, a normalization layer for training stability, and a \textit{channel mixer} for channel information interaction. The \textit{token mixer} employs a multi-scale reparameterized depthwise convolution, where the medium-kernel branch consists of five different kernel patterns to mimic the central vision enhancement feature of human eyes. The normalization layer is a Batch Normalization~\cite{batch_norm} layer, and the \textit{channel mixer} comprises a MLP module consists of two $1\times1$ pointwise convolution layers with a GELU~\cite{hendrycks2016gelu} activation function in between. Additionally, the Downsampling layer is a specialized version of the MetaNeXt block with a simplified \textit{token mixer}.}
    \label{fig:architecture}
    \vspace{-15pt}
\end{figure*}

\noindent\textbf{Large Kernel CNNs}: Traditional CNNs such as AlexNet~\cite{krizhevsky2012imagenet} and GoogLeNet~\cite{szegedy2015going} favored large kernels in their early layers, but the trend shifted towards stacking $3\times3$ kernels after VGG~\cite{simonyan2014very}.
InceptionNets~\cite{szegedy2016rethinking,szegedy2017inception} decomposed $n\times n$ convolutions into sequential $1\times n$ and $n\times 1$ convolutions for efficiency.
GCN~\cite{peng2017large} and SegNeXt~\cite{guo2022segnext} increased the kernel size through a combination of $1\times k + k\times 1$ and $k\times 1 + 1\times k$ convolutions for semantic segmentation.
ConvMixer~\cite{convmixer} achieved a substantial performance improvement through $9\times9$ depthwise convolutions inspired by the global perspective of ViTs~\cite{dosovitskiy2020image} and MLP-Mixers~\cite{tolstikhin2021mlp}.
MogaNet~\cite{iclr2024MogaNet} crafted multi-scale spatial aggregation blocks with dilated convolutions to gather discriminative features.
ConvNeXt~\cite{liu2022convnet} explored modern CNN architecture with $7\times7$ depthwise convolutions, reflecting the design philosophy of Swin Transformer~\cite{liu2021swin}.
InceptionNeXt~\cite{yu2023inceptionnext} enhanced throughput and performance by decomposing large-kernel depthwise convolutions into four parallel branches.
SKNet~\cite{li2019selective} and LSKNet~\cite{Li_2023_ICCV} combined multi-branch convolutions along the channel or spatial dimension.
RepLKNet~\cite{ding2022scaling} expanded kernel size to $31\times31$ with SRP, achieving performance comparable to Swin Transformers.
Furthermore, UniRepLKNet~\cite{ding2023unireplknet} introduced four design principles for large-kernel CNNs, demonstrating universal applicability across various modalities.
SLaK~\cite{liu2022more} incorporated stripe convolutions with dynamic sparsity to scale up kernels to $51\times51$.
PeLK~\cite{chen2024pelk} investigated convolution operations with kernels expanding up to $101\times101$ in a human-like pattern.
Additionally, LargeKernel3D~\cite{chen2023largekernel3d} and ModernTCN~\cite{donghao2024moderntcn} introduced large kernel design into 3D networks and time series analysis.

There are three major differences between prior efforts and our proposed method: (1) We adopt a simple and consistent design across inner-stage blocks and downsampling layers, facilitating easier hardware acceleration and further algorithm optimization. (2) We introduce multi-scale depthwise convolution, where the large-kernel convolution is decomposed into strip convolutions for efficiency, and the reparameterized medium-kernel convolution is meticulously crafted to imitate the central focusing characteristic of human eyes. (3) We eliminate normalization layers from SRP branches to reduce memory usage during training, enabling greater feature diversity within limited resources.  

\section{Method}

\subsection{Overall Architecture}
\label{sec:architecture}

The architecture of \modelname{} is based on RepViT~\cite{wang2023repvit}, as illustrated in Figure~\ref{fig:architecture}. The macro structure follows the four-stage framework of conventional CNNs~\cite{he2016deep} and hierarchical ViTs~\cite{liu2021swin}. It begins with a stem module consisting of two $3\times3$ convolutions with a stride of $2$~\cite{wang2023repvit,li2022rethinking,xiao2021early}.
Each subsequent stage progressively enhances the semantic representation while reducing spatial dimensions. 
The micro blocks adhere to the MetaNeXt design~\cite{liu2022convnet,yu2023inceptionnext}, incorporating a \textit{token mixer} for spatial feature extraction, a \textit{channel mixer} for visual semantic interaction, a normalization layer~\cite{batch_norm} to stabilize and accelerate training, and a shortcut connection~\cite{he2016deep} to smooth the loss landscape~\cite{visualloss}.
\begin{equation}
  Y = X + \mathrm{ChannelMixer}\big(\mathrm{Norm}(\mathrm{TokenMixer}(X))\big),
\end{equation}
where $X, Y \in \mathbb{R}^{B \times C \times H \times W}$ with $B$ represents batch size, $C$ denotes channel number, and $H$ and $W$ indicate image height and width, respectively. 
$\mathrm{Norm}(\cdot)$ denotes the Batch Normalization (BN) layer~\cite{batch_norm}. 
$\mathrm{TokenMixer}(\cdot)$ operates as a \textit{chunk convolution} when maintaining the feature scale or as a \textit{copy convolution} during downsampling. 
Meanwhile, $\mathrm{ChannelMixer}(X)=\mathrm{Conv_{1\times1,\downarrow}}\big(\sigma(\mathrm{Conv_{1\times1,\uparrow}}(X))\big)
$ is a channel MLP module comprising two fully-connected layers with an activation function in between, resembling the feed-forward network in a Transformer~\cite{vaswani2017attention}. 
Here, $\sigma$ represents the GELU~\cite{hendrycks2016gelu} activate function, and $\mathrm{Conv_{1\times1,\uparrow}}$ and $\mathrm{Conv_{1\times1,\downarrow}}$ stand for $1\times1$ pointwise convolutions for expanding and squeezing feature maps, respectively.

The downsampling layer between each stage is a modified version of the MetaNeXt block~\cite{liu2022convnet,yu2023inceptionnext}, where the shortcut connection bypasses the \textit{channel mixer}.
\begin{equation}
\begin{split}
  \hat{X} &= \mathrm{Norm}(\mathrm{TokenMixer}(X)), \\ 
  Y &= \hat{X} + \mathrm{ChannelMixer}(\hat{X}),
\end{split}
\end{equation}
where $\hat{X}, Y \in \mathbb{R}^{B \times 2C \times H/2 \times W/2}$. 
Additionally, an optional $1\times1$ pointwise convolution layer can be implemented to achieve customized output channels.

\subsection{Chunk convolution}

\begin{algorithm}[t]
\caption{Chunk Convolution in a PyTorch-like style}
\label{alg:chunk}
\begin{lstlisting}[style=Python]
class ChunkConv(Module):
    def __init__(self, in_channels):
        super().__init__()
        assert in_channels % 4 == 0
        hidden_channels = in_channels // 4
        self.s = RepDWConvS(hidden_channels)
        self.m = RepDWConvM(hidden_channels)
        self.l = Sequential(
            Conv2d(
                in_channels=hidden_channels,
                out_channels=hidden_channels,
                kernel_size=(1, 11),
                padding=(0, 5),
                groups=hidden_channel
            ),
            Conv2d(
                in_channels=hidden_channels,
                out_channels=hidden_channels,
                kernel_size=(11, 1),
                padding=(5, 0),
                groups=hidden_channels
            )
        )

    def forward(self, x):
        i, s, m, l = chunk(x, chunks=4, dim=1)
        bs = (i, self.s(s), self.m(m), self.l(l))
        return cat(bs, dim=1)
\end{lstlisting}
\end{algorithm}

\begin{table}[t]
\caption{\textbf{Complexity of different types of convolution.} The measurement is simplified by assuming consistent input and output channels and omitting the bias term. $k$, $C$, $H$ and $W$ denote kernel size, channel number, image height and width, respectively.}
\label{tab:complexity}
  \centering
  \small
  \renewcommand{\arraystretch}{1.1}
  \begin{tabular}{cccccc}
  \toprule
  \multirow{1}{*}{Convolution}          &  \multirow{1}{*}{\makecell{Parameters}} & \multirow{1}{*}{FLOPs} \\
  \hline
  Standard & $k^2C^2$ & $k^2C^2HW$\\
  Depthwise & $k^2C$ & $k^2CHW$\\
  Chunk & $(9+k^2+22)C/4$ & $(9+k^2+22)CHW/4$ \\
  \bottomrule
  \end{tabular}
  \vspace{-10pt}
\end{table}

Chunk convolution, as illustrated in Algorithm \ref{alg:chunk}, represents a specialized form of the inception depthwise convolution~\cite{yu2023inceptionnext} where each group possesses an equal number of channels for simplicity:
1. \textbf{Identity mapping}, preserving original information while reducing computation; 
2. \textbf{Repamaterized small-kernel depthwise convolution}, capturing local features and accelerating processing; 
3. \textbf{Repamaterized medium-kernel depthwise convolution}, expanding the ERF and leveraging the flexibility of SRP to emulate the central focusing feature of human eyes; 
4. \textbf{Equivalent large-kernel depthwise convolution}, comprising two layers of strip convolutions, effectively capturing the global perspective while conserving computational resources.
By incorporating this multi-scale strategy, our model aims to replicate the long-range modeling capabilities observed in ViTs while maintaining the locality and efficiency of CNNs.
Specifically, for input $X$, it is evenly partitioned into four groups along the channel dimension,
\begin{equation}
X_i, X_s, X_m, X_l = \mathrm{Chunk}(X),
\end{equation}
where $\mathrm{Chunk}(\cdot)$ splits input $X$ evenly along the channel dimension ($X_i,X_s,X_m,X_l \in \mathbb{R}^{B \times C/4 \times H \times W}$).
Next, each inputs are fed into different parallel branches,
\begin{equation}
\begin{split}
Y_i \; &= X_i, \\
Y_s \; &= \mathrm{RepDWConvS \;\,}_{k_s \; \times k_s \;} (X_s \;), \\
Y_m &= \mathrm{RepDWConvM}_{k_m \times k_m}(X_m), \\
Y_l \; &= \mathrm{DWConv}_{k_l\times 1}\big(\mathrm{DWConv}_{1\times k_l}(X_l)\big), \\
\end{split}
\end{equation}
where $k_s$ denotes the small square kernel size which is defaulted to $3$,  $k_m$ represents the medium square kernel size with a default value of $7$, and $k_l$ refers to the strip kernel size set as $11$ by default. 
$\mathrm{RepDWConvS}$ and $\mathrm{RepDWConvM}$ stand for reparameterized small and medium-kernel depthwise convolutions, respectively. 
Ultimately, the outputs from each branch are concatenated along the channel dimension,
\begin{equation} \label{eq1}
Y = \mathrm{Concat}(Y_i, Y_s, Y_m, Y_l)
\end{equation}

Specifically, $\mathrm{RepDWConvS}$ and $\mathrm{RepDWConvM}$ consist of multiple branches during training, as illustrated in Figure~\ref{fig:architecture}, which are consolidated into a single branch during inference. 
Additionally, inspired by the Peripheral Convolution~\cite{chen2024pelk} and Decomposed Manhattan Self-Attention~\cite{fan2023rmt}, we have meticulously designed $\mathrm{RepDWConvM}$ with five different kernel patterns to emulate the central focusing property of human eyes. 
\begin{equation}
\begin{split}
  Y_s\; &= \mathrm{DWConv}_{3\times3}(X_s) + \\ 
  &\;\;\;\;\; \mathrm{DWConv}_{1\times3}(X_s) + \\ 
  &\;\;\;\;\; \mathrm{DWConv}_{3\times1}(X_s) + \\ 
  &\;\;\;\;\; \mathrm{DWConv}_{2\times2,d=2}(X_s) \\
  Y_m &= \mathrm{DWConv}_{7\times7}(X_m) + \\ 
  &\;\;\;\;\; \mathrm{DWConv}_{3\times5}(X_m) + \\ 
  &\;\;\;\;\; \mathrm{DWConv}_{5\times3}(X_m) + \\
  &\;\;\;\;\; \mathrm{DWConv}_{7\times 1}\big(\mathrm{DWConv}_{1\times 7}(X_m)\big) + \\ 
  &\;\;\;\;\; \mathrm{DWConv}_{5\times 1}\big(\mathrm{DWConv}_{1\times 5}(X_m)\big)
\end{split}
\end{equation}

\begin{table*}[t]
  \caption{\textbf{Classification performance on ImgeNet-1K.} Following~\cite{li2022efficientformer,wang2023repvit}, latency is measured on an iPhone 12 with models compiled by Core ML Tools, reporting both the mean and standard deviation across 20 experimental trials. Similar to~\cite{Graham_2021_ICCV}, throughput is tested on a Nvidia RTX3090 GPU with maximum power-of-two batch size that fits in memory. ``$\dagger$'' denotes the evaluation image size is 256.}
  \label{tab:with_distillation}
  \centering
  \small
  \scalebox{0.95}{\begin{tabular}{cccccccccc} \toprule
\multirow{2}{*}{Model}           & \multirow{2}{*}{Type}      & \multirow{2}{*}{Params (M)} & \multirow{2}{*}{GMACs} &  \multirow{2}{*}{\makecell{Latency $\downarrow$ \\ (ms)}} & \multirow{2}{*}{\makecell{Throughput $\uparrow$ \\ (im/s)}} & \multirow{2}{*}{Top-1 (\%)}  \\
\\
\hline
MobileViG-Ti~\cite{munir2023mobilevig}  &   CNN-GNN   & 5.2      &  0.7    & \valstd{1.27}{0.02}   & 4337   & 75.7    \\
SwiftFormer-XS~\cite{Shaker_2023_ICCV} & Hybrid & 3.5 & 0.6 & \valstd{1.00}{0.04}  & 4304  & 75.7 \\
EfficientFormerV2-S0~\cite{li2022rethinking}  &   Hybrid   & 3.5      &  0.4    & \valstd{0.91}{0.01}      & 1274  & 75.7   \\
FastViT-T8\textsuperscript{$\dagger$}~\cite{vasu2023fastvit} & Hybrid & 3.6 & 0.7 & \valstd{0.89}{0.01} & 3909  & 76.7 \\
RepViT-M0.9~\cite{wang2023repvit} & CONV & 5.1 & 0.8 & \valstd{0.89}{0.01} & 4817  & 78.7  \\
EfficientFormerV2-S1~\cite{li2022rethinking}  &   Hybrid   & 6.1      &  0.7    & \valstd{1.06}{0.01}     & 1153   & 79.0  \\
RepViT-M1.0~\cite{wang2023repvit} & CONV & 6.8 & 1.1 & \valstd{1.02}{0.01} & 3910  & 80.0  \\
\rowcolor[HTML]{ECF4FF}
\textbf{\modelname{}-M1} & CONV & 4.8 & 0.8 & \valstdb{0.86}{0.03} & 3885   & \textbf{78.8}  \\
\rowcolor[HTML]{DAE8FC}
\textbf{\modelname{}-M2} & CONV & 6.5 & 1.1 & \valstdb{1.00}{0.04} & 3198 & \textbf{80.1}  \\
\hline
MobileViG-S~\cite{munir2023mobilevig}  &   CNN-GNN   & 7.2      &  1.0    & \valstd{1.50}{0.01}      & 2985   & 78.2    \\
SwiftFormer-S~\cite{Shaker_2023_ICCV} & Hybrid & 6.1 & 1.0 & \valstd{1.16}{0.04}  & 3376  & 78.5 \\
EfficientFormer-L1~\cite{li2022efficientformer}       & Hybrid    &    12.3   & 1.3    & \valstd{1.42}{0.02}   & 3360  & 79.2    \\
FastViT-T12\textsuperscript{$\dagger$}~\cite{vasu2023fastvit} & Hybrid & 6.8 & 1.4 & \valstd{1.33}{0.03}  & 3182  & 80.3 \\
  RepViT-M1.1~\cite{wang2023repvit} & CONV & 8.2 & 1.3 & \valstd{1.13}{0.01} & 3604  & 80.7 \\
\rowcolor[HTML]{DAE8FC}
  \textbf{\modelname{}-M3} & CONV & 7.8 & 1.3 & \valstdb{1.11}{0.04} & 2903  & \textbf{80.7}  \\
\hline
MobileViG-M~\cite{munir2023mobilevig}  &   CNN-GNN   & 14.0      &  1.5    & \valstd{1.86}{0.02}      & 2491   & 80.6    \\
FastViT-S12\textsuperscript{$\dagger$}~\cite{vasu2023fastvit} & Hybrid & 8.8 & 1.8 & \valstd{1.51}{0.03}  & 2313  & 80.9 \\
SwiftFormer-L1~\cite{Shaker_2023_ICCV} & Hybrid & 12.1 & 1.6 & \valstd{1.62}{0.02}  & 2576  & 80.9 \\
EfficientFormerV2-S2~\cite{li2022rethinking}  &   Hybrid   &  12.6  &  1.3    & \valstd{1.63}{0.01}     & 611  & 81.6  \\
FastViT-SA12\textsuperscript{$\dagger$}~\cite{vasu2023fastvit} & Hybrid & 10.9 & 1.9 & \valstd{1.66}{0.01}   & 2181  & 81.9 \\
RepViT-M1.5~\cite{wang2023repvit} & CONV & 14.0 & 2.3 & \valstd{1.51}{0.02} & 2151  & 82.3 \\
\rowcolor[HTML]{DAE8FC}
\textbf{\modelname{}-M4} & CONV & 13.3 & 2.3 & \valstdb{1.48}{0.04}  & 1745  & \textbf{82.3} \\
\hline
EfficientFormer-L3~\cite{li2022efficientformer}  &   Hybrid   & 31.3   &  3.9    & \valstd{2.79}{0.02}     & 1422  & 82.4  \\
MobileViG-B~\cite{munir2023mobilevig} & CNN-GNN & 26.7 & 2.8 &  \valstd{2.87}{0.04} & 1446  & 82.6 \\
SwiftFormer-L3~\cite{Shaker_2023_ICCV} & Hybrid & 28.5 & 4.0 & \valstd{2.99}{0.08}  & 1474  & 83.0 \\
EfficientFormer-L7~\cite{li2022efficientformer}  &   Hybrid   & 82.1   &  10.2    & \valstd{6.80}{0.02}  & 619  & 83.3  \\
EfficientFormerV2-L~\cite{li2022rethinking}  &   Hybrid   & 26.1   &  2.6    & \valstd{2.75}{0.01}     & 399  & 83.3  \\
RepViT-M2.3~\cite{wang2023repvit} & CONV & 22.9 & 4.5 & \valstd{2.24}{0.01} & 1184  & 83.3  \\
  FastViT-SA24\textsuperscript{$\dagger$}~\cite{vasu2023fastvit} & Hybrid & 20.6 & 3.8 & \valstd{2.78}{0.01}    & 1128  & \textbf{83.4} \\
\rowcolor[HTML]{DAE8FC}
\textbf{\modelname{}-M5} & CONV & 21.7 & 4.5 & \valstdb{2.20}{0.02} & 978   & 83.3  \\
\bottomrule
\end{tabular}
}
  \vspace{-10pt}
\end{table*}

The inference complexity of three types of convolution is shown in Table~\ref{tab:complexity}. 
The computational cost of chunk convolution reflects the mixed nature of the operations performed within each branch. 
By distributing the operations, chunk convolution strikes a balance between computational complexity and representational capability.

\subsection{Copy convolution}

Copy convolution as shown in Algorithm \ref{alg:copy}, is a variation of the chunk convolution, where each group operate on the same input with a stride of $2$ to reduce spatial dimensions. 
The distinction lies in the sequential stacking of strip convolutions $\mathrm{DWConv}_{3\times 1}\big(\mathrm{DWConv}_{1\times 3}(X_s)\big)$, rather than parallel execution $\mathrm{DWConv}_{3\times 1}(X_s) + \mathrm{DWConv}_{1\times 3}(X_s)$.
\begin{equation}
\begin{split}
Y_s \; &= \mathrm{RepDWConvS \;\,}_{k_s \; \times k_s \;,s=2} (X_s \;), \\
Y_m &= \mathrm{RepDWConvM}_{k_m \times k_m,s=2}(X_m), \\
\end{split}
\end{equation}
similarly, the outputs from each branch are concatenated,
\begin{equation} \label{eq1}
Y = \mathrm{Concat}(Y_s, Y_m)
\end{equation}

Additionally, a pointwise convolution can be utilized to adjust the channel dimension, providing greater flexibility.
\begin{algorithm}[t]
\caption{Copy Convolution in a PyTorch-like style}
\label{alg:copy}
\begin{lstlisting}[style=Python]
class CopyConv(Module):
    def __init__(self, in_channels):
        super().__init__()
        self.s = RepDWConvS(in_channels, stride=2)
        self.m = RepDWConvM(in_channels, stride=2)

    def forward(self, x):
        return cat((self.s(x), self.m(x)), dim=1)
\end{lstlisting}
\end{algorithm}

\section{Experiments}

We demonstrate \modelname{}’s applicability and effectiveness by conducting experiments across different vision tasks: classification on ImageNet-1K~\cite{deng2009imagenet}, object detection and instance segmentation on MS-COCO 2017~\cite{lin2014microsoft}, and semantic segmentation on ADE20K~\cite{zhou2017scene}. 
Following~\cite{li2022efficientformer,li2022rethinking,vasu2023mobileone,mehta2022separable,wang2023repvit}, we export the model using Core ML Tools and evaluate its latency on an iPhone 12 running iOS 16 utilizing the Xcode performance tool. 
Furthermore, we provide throughput analysis on a Nvidia RTX3090 GPU, adhering to the procedure in~\cite{wang2023repvit}, where we measure the throughput using the maximum power-of-two batch size that fits in memory.

\subsection{Image Classification}

\paragraph{Implementation details.} We perform image classification experiments on ImageNet-1K, employing a standard image size of 224$\times$224 for both training and testing. 
This dataset comprises approximately 1.3M training, 50k validation and 100k test images, distributed across 1000 categories.
We train all models from scratch for 300 epochs using the same training recipe as in~\cite{vasu2023fastvit,li2022rethinking,wang2023repvit}, except for the \modelname{}-M5 model, which used a weight decay of 0.03 instead of 0.025.
To ensure fair comparisons, we utilize the RegNetY-16GF~\cite{radosavovic2020designing} model with a top-1 accuracy of 82.9\% as the teacher model for distillation. 
Latency measurements are conducted on an iPhone 12 with models compiled by Core ML Tools under a batch size of 1 across 20 experimental trials.
Following~\cite{vasu2023fastvit,li2022rethinking}, we report the performance with and without distillation in~\ref{tab:with_distillation} and~\ref{tab:without_distillation}, respectively.

\begin{table}[t]
  \caption{Results without distillation on ImageNet-1K, where``$\dagger$'' denotes the evaluation image size is 256.}
  \label{tab:without_distillation}
  \centering
  \small
  \scalebox{0.85}{  \begin{tabular}{cccccccccc}
  \toprule
  \multirow{1}{*}{Model}          &  \multirow{1}{*}{\makecell{Latency (ms)}} & \multirow{1}{*}{Params(M)} & \multirow{1}{*}{Top-1 (\%)} \\
  \hline
  EfficientFormerV2-S0~\cite{li2022rethinking} & \valstd{0.91}{0.01} & 3.5 & 73.7 \\
  FastViT-T8\textsuperscript{$\dagger$}~\cite{vasu2023fastvit} & \valstd{0.89}{0.01} & 3.6 & 75.6 \\
  MobileOne-S1~\cite{vasu2023mobileone} & \valstd{0.89}{0.01}  & 4.8 & 75.9 \\
  StarNet-S3~\cite{ma2024rewrite} & \valstd{0.98}{0.01} & 3.7 & 77.4 \\
  RepViT-M0.9~\cite{wang2023repvit} & \valstd{0.89}{0.01} & 5.1 & 77.4 \\
  EfficientFormerV2-S1~\cite{li2022rethinking} & \valstd{1.06}{0.01} & 6.1 & 77.9 \\
  RepViT-M1.0~\cite{wang2023repvit} & \valstd{1.02}{0.01} & 6.8 & 78.6 \\
  \rowcolor[HTML]{ECF4FF}
  \textbf{\modelname{}-M1} & \valstdb{0.86}{0.03} & 4.8 & \textbf{77.5} \\
  \rowcolor[HTML]{DAE8FC}
  \textbf{\modelname{}-M2} & \valstdb{1.00}{0.04} & 6.5 & \textbf{78.9} \\
  \hline
  MobileOne-S2~\cite{vasu2023mobileone} & \valstd{1.14}{0.01} & 7.8 & 77.4 \\
  MobileOne-S3~\cite{vasu2023mobileone} & \valstd{1.31}{0.01} & 10.1 & 78.1 \\
  StarNet-S4~\cite{ma2024rewrite} & \valstd{1.11}{0.01} & 7.5 & 78.4 \\
  FastViT-T12\textsuperscript{$\dagger$}~\cite{vasu2023fastvit} & \valstd{1.33}{0.03} & 6.8 & 79.1 \\
    RepViT-M1.1~\cite{wang2023repvit} & \valstd{1.13}{0.01} & 8.2 & 79.4 \\
  \rowcolor[HTML]{DAE8FC}
    \textbf{\modelname{}-M3} & \valstdb{1.11}{0.04} & 7.8 & \textbf{79.4} \\
  \hline
  MobileOne-S4~\cite{vasu2023mobileone} & \valstd{1.73}{0.01} & 14.8 & 79.4 \\
  FastViT-S12\textsuperscript{$\dagger$}~\cite{vasu2023fastvit} & \valstd{1.51}{0.03} & 8.8 & 79.8 \\
  PoolFormer-S24~\cite{yu2022metaformer} & \valstd{2.45}{0.01} & 21.0 & 80.3 \\
  EfficientFormerV2-S2~\cite{li2022rethinking} & \valstd{1.63}{0.01} & 12.6 & 80.4 \\
  FastViT-SA12\textsuperscript{$\dagger$}~\cite{vasu2023fastvit} & \valstd{1.66}{0.01} & 10.9 & 80.6 \\
  RepViT-M1.5~\cite{wang2023repvit} & \valstd{1.51}{0.01} & 14.0 & 81.2 \\
  \rowcolor[HTML]{DAE8FC}
  \textbf{\modelname{}-M4} & \valstdb{1.48}{0.04} & 13.3 & \textbf{81.2} \\
  \hline
  PoolFormer-S36~\cite{yu2022metaformer} & \valstd{3.48}{0.05} & 31.0 & 81.4 \\
  RepViT-M2.3~\cite{wang2023repvit} & \valstd{2.24}{0.01} & 22.9 & 82.5 \\
  FastViT-SA24\textsuperscript{$\dagger$}~\cite{vasu2023fastvit} & \valstd{2.78}{0.01} & 20.6 & \textbf{82.6} \\
  \rowcolor[HTML]{DAE8FC}
  \textbf{\modelname{}-M5} & \valstdb{2.20}{0.02} & 21.7 & 82.4 \\
  \bottomrule
  \end{tabular}
}
  \vspace{-10pt}
\end{table}

\vspace{-5pt}
\paragraph{Results with knowledge distillation.}
As demonstrated in~\ref{tab:with_distillation}, \modelname{} achieves an optimal balance between accuracy and latency across various model sizes. 
With similar model sizes and latency, \modelname{}-M2 outperform EfficientFormerV2-S1 by 1.1\% top-1 accuracy and exhibits higher throughput.
\modelname{}-M1 and \modelname{}-M2 consistently surpass RepViT-M0.9 and RepViT-M1.0 by 0.1\% in top-1 accuracy while maintaining lower latency and fewer parameters. 
Larger models match the top-1 accuracy of their counterparts while benefiting from further parameters reduction. 
These results highlight the effectiveness and efficiency of our design, showing that a simple multi-scale CNN can outperform sophisticated architectures or complicated operators on mobile devices.

\vspace{-5pt}
\paragraph{Results without knowledge distillation.} 
As depicted in~\ref{tab:without_distillation}, \modelname{} achieves Top-1 accuracy comparable or superior to RepViT without the use of knowledge distillation, demonstrating its strong performance independently.
Furthermore, \modelname{} strikes an optimal balance among accuracy, latency, and model size.
For instance, \modelname{}-M1 achieves a Top-1 accuracy of 77.5\%, with a latency of 0.86ms and a compact size of 4.8M parameters. 
Additionally, with a latency of 1.0ms, \modelname{}-M2 surpasses RepViT-M1.0 by 0.3\% in accuracy while having 0.3M fewer parameters. 
In the case of larger models, \modelname{}-M3 delivered a 1.0\% performance improvement over StarNet-S4, with identical latency of 1.11ms.
Meanwhile, \modelname{}-M4 matches the 81.2\% accuracy of RepViT-M1.5, but with a 0.03ms speed advantage and a reduction of 0.7M parameters. 

\subsection{Downstream Tasks}

\paragraph{Object Detection and Instance Segmentation.}
We evaluate \modelname{}'s transfer ablility on object detection and instance segmentation tasks. 
Following \cite{li2022rethinking}, we integrate \modelname{} into the Mask-RCNN framework~\cite{he2017mask} and conduct experiments on the MS-COCO 2017 dataset~\cite{lin2014microsoft}. 
As shown in \Cref{tab:coco}, \modelname{} consistently outperforms the competitors in terms of AP$^{box}$ and AP$^{mask}$ while maintaining similar latency and model sizes. 
For instance, \modelname{}-M4 outperforms RepViT-m1.5 by 1.3 AP$^{box}$ and 0.5 AP$^{mask}$ with a similar latency, and matches the AP$^{box}$ and AP$^{mask}$ of SwiftFormer-L3 but operates twice as fast.
\modelname{}-M5 achieves competitive AP$^{box}$ and AP$^{mask}$ compared to RepViT-M2.3 and EfficientFormerV2-L, which are both initialized with weights pretrained for 450 epochs on ImageNet-1K. 
These results further demonstrate the advantages of large-kernel convolution in downstream tasks, as noted in~\cite{ding2022scaling}, and highlight
 the efficacy of our multi-scale kernel design, which equivalents to a grouped large-kernel depthwise convolution with additional inductive bias and efficiency trade-offs.

\begin{table*}[h]
  \centering
  \small
  \caption{
  \textbf{Object detection and instance segmentation} were evaluated using Mask R-CNN on MS-COCO 2017, 
  while \textbf{semantic segmentation} results were obtained on ADE20K. Backbone latencies were measured on an iPhone 12 with 512$\times$512 image crops using Core ML Tools. Models marked with ``$\dagger$'' were initialized with weights pretrained for 450 epochs on ImageNet-1K.
  }
  \vspace{-6pt}
  \resizebox{0.9\linewidth}{!}{\begin{tabular}{c|c|ccc|ccc|c}
\toprule
\multirow{2}{*}{Backbone} & \multirow{2}{*}{\makecell{Latency $\downarrow$ \\ (ms)}} & \multicolumn{3}{c|}{Object Detection} & \multicolumn{3}{c|}{Instance Segmentation} & \multicolumn{1}{c}{Semantic}  \\ \cline{3-9}
                          &                         & AP$^{box}$    & AP$^{box}_{50}$   & AP$^{box}_{75}$   & AP$^{mask}$    & AP$^{mask}_{50}$   & AP$^{mask}_{75}$   & mIoU   \\
                          \hline
                          \hline
ResNet18~\cite{he2016deep}                  &     4.4        & 34.0   & 54.0    & 36.7    & 31.2   & 51.0    & 32.7   & 32.9   \\
PoolFormer-S12~\cite{yu2022metaformer}            &     7.5          & 37.3   & 59.0    & 40.1    & 34.6   & 55.8    & 36.9     & 37.2  \\
EfficientFormer-L1~\cite{li2022efficientformer}        &     5.4              & 37.9   & 60.3    & 41.0    & 35.4   & 57.3    & 37.3   &  38.9  \\
FastViT-SA12~\cite{vasu2023fastvit}     &     \text{5.6}              &  \text{38.9}  &  \text{60.5}   & \text{42.2}  &    \text{35.9}    &   \text{57.6}      &  \text{38.1}   & \text{38.0}  \\
RepViT-M1.1~\cite{wang2023repvit}     &     \text{4.9}              &  \text{39.8}  &  \text{61.9}   & \text{43.5}  &    \text{37.2}    &   \text{58.8}      &  \text{40.1}   & \text{40.6}  \\
\rowcolor[HTML]{DAE8FC}
  \modelname{}-M3     &     \text{5.1}              &  \textbf{40.8}  &  \textbf{62.4}   & \textbf{44.7}  &    \textbf{37.8}    &   \textbf{59.5}      &  \textbf{40.6}   & \textbf{40.6}  \\
PoolFormer-S24~\cite{yu2022metaformer}            &     12.3            & 40.1   & 62.2    & 43.4    & 37.0   & 59.1    & 39.6  &  40.3 \\
PVT-Small~\cite{wang2021pyramid} & 53.7  & 40.4 & 62.9 & 43.8 & 37.8 & 60.1 & 40.3 & 39.8 \\
SwiftFormer-L1~\cite{Shaker_2023_ICCV}     &     \text{8.4}              &  \text{41.2}  &  \text{63.2}   & \text{44.8}  &    \text{38.1}    &   \text{60.2}      &  \text{40.7}   & \text{41.4}  \\
EfficientFormer-L3~\cite{li2022efficientformer}        &         12.4     & 41.4   & \text{63.9}    & 44.7    & 38.1   & \text{61.0}    & 40.4   &  43.5 \\
RepViT-M1.5~\cite{wang2023repvit}      &       \text{6.4}      & \text{41.6}   &  63.2  &  \text{45.3}   &  \text{38.6}   &  60.5   & \text{41.5}     &   \textbf{43.6}   \\
FastViT-SA24~\cite{vasu2023fastvit}     &     \text{9.3}              &  \text{42.0}  &  \text{63.5}   & \text{45.8}  &    \text{38.0}    &   \text{60.5}      &  \text{40.5}   & \text{41.0}  \\
\rowcolor[HTML]{DAE8FC}
  \modelname{}-M4      &       \text{6.6}      & \textbf{42.9}   &  \textbf{64.4}  &  \textbf{47.2}   &  \textbf{39.1}   &  \textbf{61.7}   & \textbf{41.7}     &   \text{43.3}   \\
SwiftFormer-L3~\cite{Shaker_2023_ICCV}     &     \text{12.5}              &  \text{42.7}  &  \text{64.4}   & \text{46.7}  &    \text{39.1}    &   \text{61.7}      &  \text{41.8}   & \text{43.9}  \\
EfficientFormerV2-S2\textsuperscript{$\dagger$}~\cite{li2022rethinking}        &         12.0     & 43.4   & 65.4    & 47.5    & 39.5   & 62.4    & 42.2   &  42.4 \\
FastViT-SA36~\cite{vasu2023fastvit}     &     \text{12.9}              &  \text{43.8}  &  \text{65.1}   & \text{47.9}  &    \text{39.4}    &   \text{62.0}      &  \text{42.3}   & \text{42.9}  \\
  EfficientFormerV2-L\textsuperscript{$\dagger$}~\cite{li2022rethinking}        &        18.2     & \text{44.7}   & \textbf{66.3}    & \text{48.8}   & 40.4   & 63.5   & 43.2   &  45.2 \\
RepViT-M2.3\textsuperscript{$\dagger$}~\cite{wang2023repvit}      &       \text{9.9}      & 44.6   &  66.1  &  \text{48.8}   &  \textbf{40.8}   &  \textbf{63.6}  & \textbf{43.9}     &   \textbf{46.1}   \\
\rowcolor[HTML]{DAE8FC}
  \modelname{}-M5      &       \text{10.4}      & \textbf{44.7}   &  66.0  &  \textbf{49.2}   &  \text{40.7}   &  \text{63.5}  & \text{43.6}     &   \text{45.0}   \\
\bottomrule
\end{tabular}
}
  \vspace{-5pt}
  \label{tab:coco}
\end{table*}

\vspace{-5pt}
\paragraph{Semantic Segmentation.}
We perform semantic segmentation experiments on the ADE20K dataset~\cite{zhou2017scene}, which consists of approximately 20K training images and 2K validation images across 150 categories.
We adhered to the training protocol from the previous works~\cite{li2022efficientformer,li2022rethinking} using the Semantic FPN framework~\cite{kirillov2019panoptic}. 
As illustrated in~\Cref{tab:coco}, \modelname{} demonstrates favorable mIoU-latency trade-offs across various model sizes. 

\begin{table*}[h]
\caption{\textbf{Ablation study conducted under 120 epochs on the ImageNet-1K classification benchmark}, using \modelname{}-M1 as the baseline. Metrics reported include Top-1 accuracy on the validation set, latency on an iPhone 12, and throughput on a RTX3090 GPU.}
  \vspace{-6pt}
\centering
\small
\scalebox{0.95}{\begin{tabular}{l|l|c c c c c } \toprule
  \multirow{2}{*}{Ablation} & \multirow{2}{*}{Variant} & \multirow{2}{*}{\makecell{Params \\ (M)}} & \multirow{2}{*}{\makecell{Latency \\ (ms)}} & \multirow{2}{*}{\makecell{Throughput \\ (im/s)}} & \multirow{2}{*}{\makecell{Top-1 \\ (\%)}} \\
  ~ & ~ & ~ & ~ & ~ & ~ \\
\hline
Baseline & None (\modelname{}-M1) & 4.82 & \valstd{0.86}{0.03} & 3885 & 75.34 \\
\hline
Downsample & simple $3\times3$ convolution & 4.89 & \valstd{0.87}{0.04} & 4078 & 74.45 \\
\hline
\multirow{3}{*}{\makecell[l]{Branch}} &  remove small kernel & 4.81 & \valstd{0.85}{0.03} & 4017 & 75.22 \\
~ & remove medium kernel & 4.77 & \valstd{0.85}{0.03} & 4341 & 75.25 \\
~ & remove large kernel & 4.79 & \valstd{0.84}{0.03} & 4332 & 75.22 \\
\hline
  \multirow{3}{*}{\makecell[l]{Medium kernel}}
~ & add $5\times5$ and $3\times3$ kernels & 4.82 & \valstd{0.86}{0.03} & 3885 & 75.62 \\
~ & $5\times5$ and $3\times3$ kernels $\rightarrow$ $5\times3$ and $3\times5$ kernels & 4.82 & \valstd{0.86}{0.03} & 3885 & 75.69 \\
~ & add sequential $1\times7,7\times1$ and $1\times5,5\times1$ kernels & 4.82 & \valstd{0.86}{0.03} & 3885 & 75.71 \\
\cdashline{2-6}
\multirow{2}{*}{\makecell[l]{Small kernel}}
~ & add sequential $1\times3,3\times1$ and dilated $2\times2$ kernels & 4.82 & \valstd{0.86}{0.03} & 3885 & 75.84 \\
  ~ & sequential $1\times3,3\times1$ $\rightarrow$ parallel operations & 4.82 & \valstd{0.86}{0.03} & 3885 & \textbf{75.97} \\
\hline
\multirow{2}{*}{\makecell[l]{RepViT}}
  ~ & None(RepViT-M0.9) & 5.07 & \valstd{0.89}{0.01} & 4817 & 75.19 \\
  ~ & Downsample $\rightarrow$ \modelname{}'s Downsample & 4.99 & \valstd{0.89}{0.01} & 4731 & 75.32 \\
\cdashline{2-6}
\multirow{2}{*}{\makecell[l]{ConvNeXt}}
  ~ & None(ConvNeXt-femto) & 5.22 & - & 3636 & 72.37 \\
  ~ & Downsample $\rightarrow$ \modelname{}'s design & 5.25 & - & 3544 & 74.28 \\
  \bottomrule
\end{tabular}
}
\label{tab:ablation}
  \vspace{-10pt}
\end{table*}

\subsection{Ablation Studies}
We conduct ablation studies under 120 epochs on ImageNet-1K~\cite{deng2009imagenet} using \modelname{}-M1 without SRP as baseline from the following aspects.

\textbf{Downsampling layer.} As illustrated in Table~\ref{tab:ablation}.
The baseline model serves as the reference point with a Top-1 accuracy of 75.34\%. 
This model includes the full architecture with all kernel branches and the will designed downsampling layers.
It is clear that replacing the downsampling layers with simple $3\times3$ convolutions results in a noticeable drop in Top-1 accuracy to 74.45\%, which is a decrease of 0.9\% compared to the baseline. 
This change implies that the well designed downsampling layers in the baseline architecture are crucial for maintaining higher accuracy.
Additionally, substituting RepVit’s downsampling layer with our proposed modification slightly increased accuracy from 75.19\% to 75.32\% without affecting latency. Our change to ConvNeXt also improved accuracy by 1.9\%.

\textbf{Kernel branches.} Table~\ref{tab:ablation} shows that each kernel branch contributes incrementally to the overall Top-1 accuracy of the model. 
The removal of any single branch leads to a slight decrease in accuracy. 
For example: removing the small kernel branch leads to a minor reduction in Top-1 accuracy to 75.22\%, excluding the medium kernel branch results in a Top-1 accuracy of 75.25\%, and the Top-1 accuracy drops to 75.22\% when the large kernel branch is eliminated.
This highlights the collective contribution of multi-scale kernels in improving the model's performance.

\textbf{Structural reparameterization.} Table~\ref{tab:ablation} illustrates the substantial enhancement in Top-1 accuracy by incorporating the SRP mechanism.
For the medium-kernel branches, adding $5\times5$ and $3\times3$ convolution kernels increase the accuracy from 75.34\% to 75.62\%. 
Subsequently, substituting these kernels with $5\times3$ and $3\times5$ kernels further boosts the performance to 75.69\%. 
Moreover, integrating two sequentially stacked strip convolutions into the branch slightly increases the accuracy to 75.71\%.
These incremental advancements collectively highlight the effectiveness of our design, which is specifically engineered to emulate the human foveal vision system.
Refining the small kernel branch by introducing a dilated $2\times2$ convolution and a series of concatenated strip convolutions has substantially lifted the accuracy to 75.84\%. 
Additionally, the transition from serial to parallel branches for the strip convolutions has further elevated accuracy to 75.97\%, surpassing previous records~\cite{li2022rethinking,Shaker_2023_ICCV} achieved with knowledge distillation under 300 epochs, as detailed in Table~\ref{tab:with_distillation}. 
Overall, depending on its versatility and efficacy, SRP is becoming the default option for designing lightweight network architectures~\cite{vasu2023fastvit,wang2023repvit}.

\subsection{CAM Analysis}

\begin{figure}[t]
  \centering
    \includegraphics[width=1.0\linewidth]{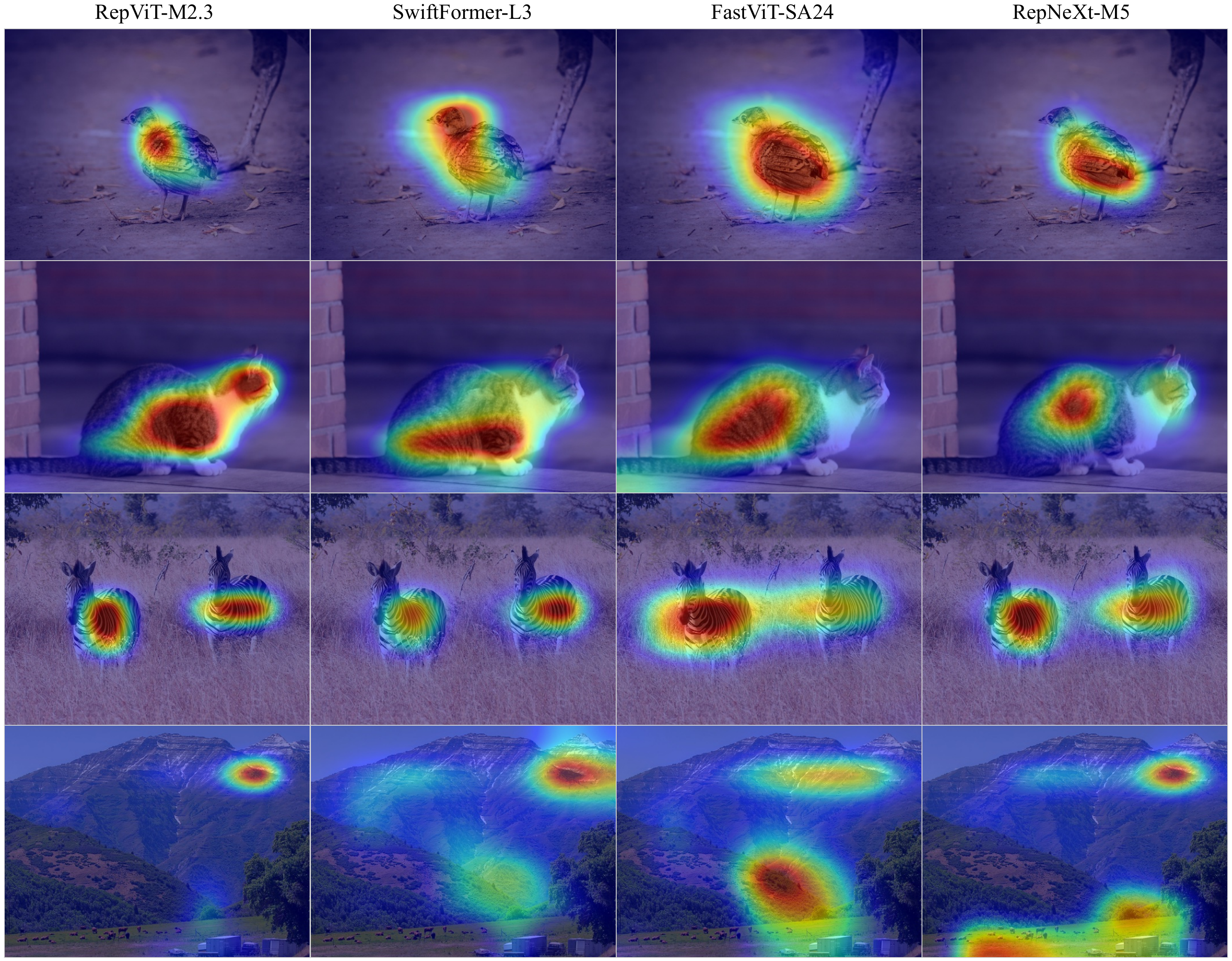}
    \caption{\textbf{Grad-CAM} on the MS-COCO validation dataset for RepViT-M2.3, SwiftFormer-L3, FastViT-SA24 and \modelname{}-M5. \modelname{} captures local details similar to RepViT while providing a global perspective comparable to FastViT.} 
    \label{fig:cam}
    \vspace{-0.5cm}
\end{figure}

We visualize class activation maps (CAM) using Grad-CAM~\cite{cvpr2017grad} with the TorchCAM Toolbox~\cite{torcham2020}.
As illustrated in Figure~\ref{fig:cam}, \modelname{} can capture local features like RepViT while also enjoying a global view akin to FastViT.

\section{Conclusions}

In this paper, we introduced a multi-scale depthwise convolution integrated with both serial and parallel SRP mechanisms, enhancing feature diversity and expanding the network’s expressive capacity without compromising inference speed.
Specifically, we designed a reparameterized medium-kernel convolution to imitate the human foveal vision system.
Additionally, we proposed our light-weight, general-purpose \modelname{}s that employed the \textit{distribute-transform-aggregate} design philosophy across inner-stage blocks as well as downsampling layers, achieving comparable or superior accuracy-efficiency trade-off across various vision benchmarks, especially on downstream tasks.
Moreover, our flexible multi-branch design functions as a grouped depthwise convolution with additional inductive bias and efficiency trade-offs.
It can also be reparameterized into a single-branch large-kernel depthwise convolution, enabling potential optimization towards different accelerators.

For future enhancements, we plan to delve into optimizations towards large kernel designs, investigate SRP upon \textit{channels mixers}, extend \modelname{} to more vision tasks and other modalities, and scale up our models further.
We hope our simple yet effective design will inspire further research towards light-weight models.

\textit{Limitations}. One limitation of \modelname{} is its marginal improvement in both accuracy, speed, and model size compared to the previous state-of-the-art (SOTA) model~\cite{wang2023repvit}. Additionally, it experiences a substantial increase in inference time when dealing with larger images due to large-kernel convolutions. We aim to address these shortcomings in future iterations.

{
    \small
    \bibliographystyle{ieeenat_fullname}
    \bibliography{main}
}


\end{document}